\documentclass{fundam}[a4]
\usepackage{url} % takes care of hyperlinks, preferred over hyperref
\usepackage{graphicx}% allows for inclusion of graphic files (figures)
\usepackage{amsmath,amsfonts}
\usepackage{amssymb}
\usepackage{latexsym}
\usepackage{wasysym}
\usepackage{amscd}
\usepackage{mathrsfs}
\date{\, }

\newcommand {\ra}{\rightarrow}

\begin{document}
\setcounter{page}{1001}
\issue{XXI~(2022)}

\title{A Note on Categories about Rough Sets}%\tnoteref{mytitlenote}}

\address{Churn-Jung Liau, email: liaucj@iis.sinica.edu.tw}

\author{Yu-Ru Syau \\ Department of Information Management, National Formosa University \\ Huwei 63201, Yunlin, Taiwan
\and
En-Bing Lin \\ Department of Mathematics, Central Michigan University \\
Mt. Pleasant, Michigan 48859, USA \and Churn-Jung Liau
\\ Institute of Information Science
\\ Academia Sinica, Nankang 115, Taipei, Taiwan}
\maketitle

\runninghead{Y. R. Syau, C. J. Liau, E. B. Lin}{Categories about Rough Sets}

\begin{abstract}
Using the concepts of category and functor, we provide some insights and prove an intrinsic property of the category ${\bf AprS}$ of approximation spaces and relation-preserving functions, the category ${\bf RCls}$ of rough closure spaces and continuous functions, and the category ${\bf RInt}$ of rough interior spaces and continuous functions. Furthermore, we define the category ${\bf IS}$ of information systems and O-A-D homomorphisms, and establish the relationship between the category ${\bf IS}$ and the category ${\bf AprS}$ by considering a subcategory ${\bf NeIS}$ of  ${\bf IS}$ whose objects are information systems and whose arrows are non-expensive O-A-D homomorphisms with surjective attribute mappings.
\end{abstract}
\begin{keywords}
Rough sets, approximation spaces, equivalence relations, topological spaces, clopen topologies.
%\MSC[2020] 03E75\sep  94D99
\end{keywords}

%\linenumbers

\section{Introduction}
Pawlak's rough set theory (RST) \cite{Pawlak1982} is intrinsically based on an equivalence relation on a domain of interests. RST starts from the notion of approximation space which is defined as a pair $\langle U,t \rangle$, where $U$ is a set and $t$ is an equivalence relation on $U$.

In an approximation space $\langle U,t \rangle$, the classical upper and lower approximations as mappings between subsets of $U$ are indeed a pair of closure and interior operators on the set $U$. Moreover, the set of fixed points of the lower approximation is a topology in which every open set is also a closed set, namely a clopen topology.

On the other hand, categories are algebraic structures with many complementary natures. Almost every known example of a mathematical structure with the appropriate structure-preserving map yields a category. Category theory has provided the foundations for many of the twentieth century's greatest advances in pure mathematics. However, to the best of our knowledge, there are not much work done in incorporating both category theory and rough set theory. Recently, Borzooei et al. \cite{Borzooei} considered the class of approximation spaces and discussed some categories of lower, upper and natural approximations without using much topological structures of spaces. It is not fruitful in obtaining more results by using limited tools in some categories.

In this paper, we define the category ${\bf AprS}$ of approximation spaces and relation-preserving function. We characterize clopen topologies on a set by introducing the concept of rough closure operator and rough interior operator on the set. We prove the equivalence of three categories, namely, the category of all rough interior spaces and continuous functions, the category of rough closure spaces and continuous functions, the category of
approximation spaces and relation-preserving functions. We organize our paper as follows. In Section 2, we provide some preliminary backgrounds, followed by some definitions of several categories and main results in Section 3. We survey and compare some categories about rough sets in Section 4 and finally conclude with some remarks in Section 5.

\section{Preliminaries}
A (binary) relation $R$ on a set $U$ is a subset of the Cartesian product $U \times U$. We write $(u,v) \in R$ and $uRv$ interchangeably.

Let $f:U\ra V$ be a function from a set $U$ into a set $V$. The inverse image of a set $Y \subseteq V$ under $f$, designated by $f^{\leftarrow}(Y)$, is the set
\begin{equation}
	\label{preimage}
	f^{\leftarrow}(Y)=\{u \in U  \mid f(u) \in Y \}.
\end{equation}
Similarly, the direct image of a set $X \subseteq U$ under $f$, designated by
$f[X]$, is the set
\begin{eqnarray}
	\label{image}
	f[X] &=& \{f(u) \mid u \in X \}  \nonumber\\
	&=&\{v \in V \mid v=f(u) \ \ \mbox{for some} \ u \in X \}.
\end{eqnarray}

\subsection{The notion of topological spaces}
A topological space is a set $U$ together with a set ${\cal T}$ of subsets of $U$ that includes $\emptyset$ and the whole set $U$, and is closed under finite intersections and arbitrary unions. The set ${\cal T}$ is called a topology on $U$, and its members are called open sets. The complement of an open set is called a closed set.

The interior ${X}^0$ of a subset $X$ of a topological space $U$ is the union of all open sets contained in $X$.  Accordingly, ${X}^0$ is open and is the largest open set contained in $X$. Dually, the closure ${\overline X}$ of $X$ is the intersection of all closed sets containing $X$. Hence ${\overline X}$ is closed and is the smallest closed set containing $X$.

A set $N$ in a topological space is a neighborhood of a point $u$ in the topological space if and only if (sometimes shortened to iff) $N$ contains an open set to which $u$ belongs \cite{Kelley}. The set ${\cal N}_u$ of all neighborhoods of $u$ is called the neighborhood system of $u$. A neighborhood of $u$ is called a minimal neighborhood of $u$ if it contains no member of ${\cal N}_u$ as proper subsets. In general, such a minimal member of ${\cal N}_u$ may or may not exist \cite{NSVPGRS}.

An Alexandroff space \cite{Arenas} is a topological space such that every point has a minimal neighborhood, or equivalently, arbitrary intersection of open sets is open.

Recall \cite{Kelley} that, by definition, a function $f$ from a topological space $U$ into a topological space $V$ is continuous iff the inverse image of each open set is open.

\begin{proposition}
	\label{Conditions}
	{\rm \cite{Kelley}} Assume that $U$ and $V$ are topological spaces, and that $f$ is a function from $U$ into $V$. The following statements are equivalent.
	\begin{enumerate}
		\item The function $f$ is continuous.
		\item The inverse image of each closed set is closed.
		\item For each $u \in U$ the inverse image of every neighborhood of $f(u)$ is a neighborhood of $u$.
		\item For each $u \in U$ and each neighborhood $M$ of $f(u)$ there is a neighborhood $N$ of $u$ such that $f[N] \subseteq M$.
		\item For each subset $X$ of $U$, $f[{\overline X}]  \subseteq {\overline {f[X]}}$.
		\item For each subset $Y$ of $V$, ${\overline {f^{\leftarrow}(Y)}}  \subseteq f^{\leftarrow}({\overline Y})$.
	\end{enumerate}
\end{proposition}

\subsection{Closure and interior spaces}
A closure operator on a set $U$ is a function $c: 2^U \longrightarrow 2^U$ from the power set $2^U$ of $U$ into itself which satisfies the Kuratowski closure axioms. Namely, (KC1) $c(\emptyset)=\emptyset$, (KC2) $X \subseteq c(X)$ (Extensiveness), (KC3) $c(c(X))=c(X)$ (Idempotency) and (KC4) $c(X \cup Y)=c(X) \cup c(Y)$ hold for any subsets $X$ and $Y$ of $U$. The pair $(U,c)$ is called a closure space. Notice that axiom (KC4) implies axiom (KC5) if $X \subseteq Y$ then $c(X) \subseteq c(Y)$, i.e., $c$ is isotone.

A closure operator $c$ on $U$ corresponds to the interior operator $i_{c}: 2^U \longrightarrow 2^U$:
\begin{equation}
	\label{dualC}
	i_{c}(X)=U-c(U-X) \ \ \ \ \mbox{for all} \ X \subseteq U,
\end{equation}
which satisfies corresponding axioms: (KI1) $i_{c}(U)=U$, (KI2) $i_{c}(X) \subseteq X $ (Contraction), (KI3) $i_{c}(i_{c}(X))=i_{c}(X)$ (Idempotency) and (KI4) $i_{c}(X \cap Y)=i_{c}(X) \cap d(Y)$.
The pair $(U,c)$ is called a closure space. As it is known  \cite{Kelley}, $c: 2^U \longrightarrow 2^U$ defines the topology
\begin{eqnarray}
	\label{FixP}
	{\cal T}_c &=& \{X \subseteq U \mid c(U-X)=U-X \}  \nonumber\\
	&=& \{ X \subseteq U  \mid i_{c}(X)=X  \}
\end{eqnarray}
on $U$, called the topology associated with the closure operator $c$. It follows from (\ref{dualC}), (KC3), and (KC5) that for any subset $X$ of $U$,
\begin{equation}
	\label{ClX}
	c(X) \ \mbox{\rm {is the smallest closed set containing}} \ X.
\end{equation}

It is easily seen \cite{Munkres} that $u \notin c(X)$ iff there exists an open set including $u$ that does not intersect $X$. Equivalently,
\begin{equation}
	\label{ClNbhd}
	u \in c(X) \ \ \Longleftrightarrow \ \ \mbox{\rm {every neighborhood of}} \  x \ \mbox{\rm {intersects}} \ X.
\end{equation}
Furthermore, for all $u,~v \in U$
\begin{eqnarray}
	\label{min}
	u \in c(\{v\})
	&\Longleftrightarrow&  u \ \mbox{\rm {is contained in all closed sets that include}} \ v \nonumber\\
	&\Longleftrightarrow&  v \ \mbox{\rm {is contained in all open sets that include}} \ u.
\end{eqnarray}

According to Proposition \ref{Conditions} and (\ref{FixP}), a function $f$ between closure spaces $(U,c_{U})$ and $(V,c_{V})$ is continuous if
\begin{center}
	$f[c_{U}(X)]  \subseteq c_{U}(f[X]) \ \ \ \ \mbox{for all} \ X \subseteq U$.
\end{center}
or equivalently, ${c_{U}({f^{\leftarrow}(Y)}}) \subseteq f^{\leftarrow}(c_{V}(Y))$ for all $Y \subseteq V$. Accordingly, a function $f$ between interior spaces $(U,i_{c_{U}})$ and $(U,i_{c_{V}})$ is continuous if
\begin{equation}
	\label{ContInt}
	f^{\leftarrow}(i_{c_{V}}(Y)) \subseteq  i_{c_{U}}(f^{\leftarrow}(Y)) \ \ \ \ \mbox{for all} \ Y \subseteq V.
\end{equation}

\subsection{Approximation spaces}
The basic construct of rough set theory is the {\em approximation space\/}, that is, a pair $\langle U,t \rangle$, where $U$ is the universe and $t\subseteq U\times U$ is an equivalence relation on $U$. In an approximation space $\langle U,t \rangle$, the upper and lower approximations, ${{\overline {apr}}_t}(X)$ and ${{\underline {apr}}_t}(X)$, of a set $X \subseteq U$ are defined respectively as follows \cite{Pawlak1982}:
\begin{equation}
	\label{Uapp}
	{{\overline {apr}}_t}(X) =  \{u \in U  \mid  [u]_t \cap X \ne \emptyset \}
\end{equation}
\begin{equation}
	\label{Lapp}
	{{\underline {apr}}_t}(X) = U-{{\overline {apr}}_t}(U-X)= \{u \in U  \mid  [u]_t \subseteq X  \}
\end{equation}
where $[u]_t$ is the $t$-equivalence class of $u \in U$. They are indeed a pair of closure and interior operators on the set $U$. Moreover, for every $X \subseteq U$ \cite{Pawlak1982}:
\begin{equation}
	\label{sym}
	{{{\overline {apr}}_t}(X)}= {{\underline {apr}}_t}({{\overline {apr}}_t}(X)) \ \ \ {\rm and} \ \ \
	{{{\underline {apr}}_t}(X)}= {{\overline {apr}}_t}({{\underline {apr}}_t}(X)).
\end{equation}
Consequently, the topology associated with the upper approximation is clopen and therefore Alexandroff. This, together with
(\ref{FixP}) and (\ref{Uapp}), leads to the following:
\begin{lemma}
	\label{RoughCL}
	For an approximation space $\langle U,t \rangle$, the topology
	\begin{eqnarray}
		\label{tau}
		{\tau}_t &=& \{X \subseteq U \mid {{\overline {apr}}_t}(U-X)=U-X \} \nonumber\\
		&=& \{X \subseteq U \mid {{\underline {apr}}_t}(X)=X \}
	\end{eqnarray}
	associated with ${{\overline {apr}}_t}: 2^U \longrightarrow 2^U$ is clopen and therefore Alexandroff. Moreover, the set of all $t$-equivalence classes is unique minimal base for ${\tau_t}$.
\end{lemma}

\subsection{Information systems}
As rough set theory is applied to data analysis, an approximation space is usually induced from {\em information systems\/}. According to \cite{Pawlak1982},  an information system\footnote{Also called knowledge representation systems, data tables, or attribute-value systems} is defined as a tuple $T=(U, A,\{D_a\mid a\in A\}, \{f_a\mid a\in A\})$, where $U$ is a nonempty finite set, called the universe;  $A$ is a
nonempty finite set of primitive attributes;  for each $a\in A$,
$D_a$ is the domain of values for $a$; and for each $a\in A$,
$f_a:U\ra D_a$ is the information function for the attribute $a$. To simplify the presentation, we can identify each attribute $a$ with its information function and set $D=\bigcup_{a\in A}D_a$. Then, an information system is simply a triplet $(U,A,D)$, where $A$ is a set of functions from $U$ to $D$. Given an information system  $(U,A,D)$ and a subset of attributes $B\subseteq A$, the {\em indiscernibility relation\/} with respect to $B$ is defined as $ind(B)=\{(x,y)\mid x,y\in U, a(x)=a(y)\forall a\in B\}.$  Obviously, for each $B\subseteq A$, $\langle U, ind(B)\rangle$ is an approximation space. In particular, we can associate with each information system  $(U,A,D)$ its finest approximation space $\langle U,t_A\rangle$, where $t_A=ind(A)$. On the other hand, given an approximation space $\langle U,t\rangle$, we can (somewhat trivially) associate with it a single-attribute information system $(U,\{a_t\},D)$, where $D=U/t$ is the set of equivalence classes (i.e.\ the quotient set) of $U$ with respect to $t$ and $a_t:U\to D, u\mapsto [u]_t$.

In \cite{GB}, homomorphisms between information systems are introduced. Given two information systems $T_1=(U_1,A_1,D_1)$ and $T_2=(U_2,A_2,D_2)$, an O-A-D homomorphism from $T_1$ to $T_2$ is a triple of function $h=(h_O,h_A,h_D)$, where
\begin{itemize}
	\item $h_O: U_1\to U_2$ is the object mapping,
	\item $h_A: A_1\to A_2$ is the attribute mapping, and
	\item $h_D: D_1\to D_2$ is the domain mapping,
\end{itemize}
such that \begin{equation}\label{oad}h_D(a(x))=h_A(a)(h_O(x))\end{equation}
for any $a\in A_1$ and $x\in U_1$.

\subsection{Categories and functors}
A category ${\bf C}$ is a class of objects $A, B, C,\ldots$ together with a family of disjoint sets ${\rm hom}_{\bf C}(A,B)$, one for each ordered pair $A, B$ of objects. Write $f : A \to B$ for $f \in {\rm hom}_{\bf C}(A,B)$, and call $f$ an arrow of ${\bf C}$ with source $A$ and target $B$. Given each pair $f : A \to B$, $g : B \to C$ of arrows, there is unique arrow
$gf=g \circ f : A \to C$, called the composite of $f$ and $g$. The composition is associative, and each object has an identity arrow  that serves as a unit under composition \cite{MacLane}.

A subcategory of ${\bf C}$ is a category ${\bf D}$ whose objects are objects of ${\bf C}$ and whose arrows are arrows of ${\bf C}$ with the same identities and composition of arrows.

An arrow $f : A \to B$ in a category ${\bf C}$ is called an isomorphism, and $A$ and $B$ are said to be isomorphic in ${\bf C}$, if and only if it has an inverse arrow, i.e., if there is and only if an arrow $g: B \to A$ with $g \circ f={\bf 1}_{A}$ and $f \circ g={\bf 1}_{B}$. Also, $f : A \to B$ is called a monomorphism (or monic) if for any $g,h:C\to A$, $f\circ g=f\circ h$ implies $g=h$, and an epimorphism (or epic) if for any $g,h:B\to C$, $g\circ f=h\circ f$ implies $g=h$.

In what follows, a full subcategory ${\bf D}$ of ${\bf C}$ is a subcategory of ${\bf C}$ such that
\begin{equation}
	\label{CondFulSub}
	{\rm hom}_{\bf D}(A,B)={\rm hom}_{\bf C}(A,B) \ \ \ \ \ ~\mbox{for all objects} \ A \ {\rm and} \ B \ of \ {\bf D}
\end{equation}
A small category is a category whose objects, and hence all of those arrows, form a set.

Following are some examples of small categories:

\vskip .05in
$\bullet$~The category ${\bf Set}$ of sets and functions has objects including all sets $A, B, \cdots$, and arrows of all (total) functions from $A$ into $B$ with the usual composition.

\vskip .05in
$\bullet$~The category ${\bf Top}$ of topological spaces and continuous functions.

\vskip .05in
$\bullet$~The category ${\bf Alex}$ of Alexandroff topological spaces and continuous functions.

\vskip .05in
$\bullet$~The category ${\bf Clop}$ of clopen topological spaces and continuous functions.

\vskip .05in
$\bullet$~The category ${\bf Cls}$ of closure spaces and continuous functions \cite{Seal}.

\vskip .05in
$\bullet$~The category ${\bf Int}$ of interior spaces  and continuous functions \cite{Seal}.

\vskip .05in
\noindent
The category ${\bf Clop}$ is a full subcategory of ${\bf Alex}$. The categories ${\bf Alex}$ and ${\bf Clop}$ are full subcategories of the category ${\bf Top}$.

\begin{definition}
	\label{functor}
	A functor $F : {\bf C} \to {\bf D}$ from a category ${\bf C}$ to a category ${\bf D}$ consists of an object function $F$ and an arrow function, also written $F$, in such a way that
	
	\vskip .05in
	$\bullet~F(f : A \to B)=F(f): F(A) \to F(B)$,
	
	\vskip .05in
	$\bullet~F({\bf 1}_{A})={\bf 1}_{F(A)}$, \ {\rm and}
	
	\vskip .05in
	$\bullet~F(g \circ f)=F(g) \circ F(f)$, for each pair $f : A \to B$, $g : B \to C$ of arrows of ${\bf C}$.
\end{definition}
Under this composition, a functor can be viewed as an arrow of categories. In particular, the identity function on objects and arrows in a category ${\bf C}$ is the identity functor ${{\bf 1}_{\bf C}}$. Thus, we can define the category ${\bf Cat}$ of all small categories, whose arrows are the functors between categories. This allows us to define isomorphic categories, i.e., two categories are isomorphic if they are isomorphic as objects of ${\bf Cat}$.

\section{Categories of Clopen Topologies}
In this section, we define the category ${\bf AprS}$ of approximation spaces and relation-preserving functions, and show that the relation-preserving functions between approximation spaces are precisely the continuous functions between the corresponding clopen topological spaces. We characterize clopen topologies on a set by introducing the concept of rough closure operator and rough interior operator on the set, and define the category ${\bf RCls}$ of rough closure spaces and continuous functions and the category ${\bf RInt}$ of rough interior spaces and continuous functions. We prove the equivalence of ${\bf AprS}$, ${\bf RCls}$ and ${\bf RInt}$. Furthermore, we define the category ${\bf IS}$ of information systems and O-A-D homomorphisms. We establish the relationship between ${\bf IS}$ and  ${\bf AprS}$ by considering a subcategory ${\bf NeIS}$ of information systems non-expansive O-A-D homomorphisms.

\subsection{The category of approximation spaces}
In \cite{CPCT}, Rydeheard and Burstall considered the category ${\bf Rel}$ with objects ordered pairs $(U,R)$, where $U$ is a set and $R$ is a binary relation on $U$. Arrows $f : (U,R) \to (V,S)$ are relation-preserving functions; i.e., functions $f : U \to V$ such that
$uRu'$ implies $f(u)Sf(u')$.

Considering the category ${\bf Rel}$, we can define a second category ${\bf AprS}$ of approximation spaces and relation-preserving functions, which is the subcategory of ${\bf Rel}$ with objects ordered being pairs $\langle U,t \rangle$, where $U$ is a set and $t$ is an equivalence relation on $U$, and arrows $f : \langle U,t \rangle \to \langle V,s \rangle$ being relation-preserving functions; i.e., functions $f : U \to V$ such that
\begin{equation}
	\label{Rpreserving}
	(u,u') \in t \ \ \ {\rm implies} \ \ \  (f(u),f(u')) \in s.
\end{equation}
When thought of as a relation-preserving function from $\langle U,t \rangle$ to $\langle V,s \rangle$, we will symbolize the function as $f : \langle U,t \rangle \to \langle V,s \rangle$.

Consider two approximation spaces $\langle U,t \rangle$ and  $\langle V,s \rangle$, we have the two sets of arrows
\begin{center}
	${\rm hom}_{\bf Rel}(\langle U,t \rangle,\langle V,s \rangle) \ \ \ \mbox{and} \ \ \ {\rm hom}_{\bf AprS}(\langle U,t \rangle,\langle V,s \rangle)$
\end{center}
in the categories. As functions sets of functions, these two sets are the same. This shows that for each ordered pair $\langle U,t \rangle$, $\langle V,s \rangle$ of objects of ${\bf AprS}$,
\begin{equation}
	\label{RelAprS}
	{\rm hom}_{\bf Rel}(\langle U,t \rangle,\langle V,s \rangle)={\rm hom}_{\bf AprS}(\langle U,t \rangle,\langle V,s \rangle).
\end{equation}
Thus, ${\bf AprS}$ is a full subcategory of ${\bf Rel}$.

\begin{lemma}
	\label{base} Let $\langle U,t \rangle$ and  $\langle V,s \rangle$ be approximation spaces. A function $f : \langle U,t \rangle \to \langle V,s \rangle$ is relation-preserving if and only if
	\begin{equation}
		\label{Eclass}
		f[[u]_t]  \subseteq  [f(u)]_s \ \ \ \ ~\mbox{for each} \ \ u \in U,
	\end{equation}
\end{lemma}
\noindent
{\bf Proof} \ Assume that $f : \langle U,t \rangle \to \langle V,s \rangle$ is a relation-preserving function. Let $u \in U$. If $v \in f[[u]_t]$ then, from (\ref{image}), we have $v = f(u')$ for some $u' \in [u]_t$ and therefore, from (\ref{Rpreserving}), $(f(u),f(u')) \in s$ which is equivalent to $v = f(u') \in [f(u)]_s$. This gives (\ref{Eclass}).

Conversely, assume that (\ref{Eclass}) holds. Let $(u,u') \in t$. Then $[u]_t=[u']_t$ and therefore $f[[u]_t]=f[[u']_t]$. By (\ref{Eclass}), $f[[u]_t]=f[[u']_t]  \subseteq  [f(u)]_s$. It follows that $f(u) \in [f(u)]_s$ and $f(u) \in [f(u)']_s$; hence $(f(u),f(u')) \in s$. This gives (\ref{Rpreserving}). This establishes Lemma \ref{base}.

\vskip .1in
Since the set of all equivalence classes is unique minimal base for the clopen topology associated with ${{\overline {apr}}_t}: 2^U \longrightarrow 2^U$, it follows from Item 4 of Proposition \ref{Conditions} that a function $f : \langle U,t \rangle \to \langle V,s \rangle$ is continuous iff (\ref{Eclass}) holds. This observation, together with Proposition\ref{Conditions}, Lemma \ref{RoughCL}, Lemma \ref{base} and (\ref{ContInt}), leads to the following:

\begin{proposition}
	\label{RelTop}
	Assume that $\langle U,t \rangle$ and $\langle V,s \rangle$ are approximation spaces, and that $f$ is a function from $U$ into $V$. The following statements are equivalent.
	\begin{enumerate}
		\item The function $f$ between approximation spaces is relation-preserving.
		\item The function $f$ between clopen spaces associated with the upper (resp.\ lower) approximations is continuous.
		\item For each $u \in U$, $f[[u]_t]  \subseteq  [f(u)]_s$.
		\item For each subset $X$ of $U$, $f[{{\overline {apr}}_t}(X)]  \subseteq  {{\overline {apr}}_s}(f[X])$.
		\item For each subset $Y$ of $V$, ${{\overline {apr}}_t}(f^{\leftarrow}(Y))  \subseteq f^{\leftarrow}({{\overline {apr}}_s}(Y))$.
		\item For each subset $Y$ of $V$, $f^{\leftarrow}({{\underline {apr}}_s}(Y)) \subseteq  {{\underline {apr}}_t}(f^{\leftarrow}(Y))$.
	\end{enumerate}
\end{proposition}

\vskip .1in
Since homeomorphisms are the isomorphisms in ${\bf Top}$, it follows from Proposition \ref{RelTop} that an arrow $f : \langle U,t \rangle \to \langle V,s \rangle$ is an isomorphism if and only if the function $f : U \to V$ is bijective and its inverse function $f^{-1}: V \to U$ is relation-preserving. Therefore, from Lemma \ref{base}, we obtain the following:

\begin{proposition}
	\label{IsoAprS}
	An arrow $f : \langle U,t \rangle \to \langle V,s \rangle$ in ${\bf AprS}$ is an isomorphism if and only if
	\begin{center}
		$f[[u]_t]=[f(u)]_s \ \ \ \ ~\mbox{for each} \ \ u \in U.$
	\end{center}
\end{proposition}

\subsection{The categories of rough closure and rough interior spaces}
Note that (\ref{Lapp}) and (\ref{sym}) may be written as equivalent conditions
\begin{equation}
	\label{symU}
	{{\overline {apr}}_t}(X)= U-{{{\overline {apr}}_t}(U-{{{\overline {apr}}_t}(X)})},
\end{equation}
or
\begin{equation}
	\label{symL}
	{{\underline {apr}}_t}(X)= U-{{{\underline {apr}}_t}(U-{{{\underline {apr}}_t}(X)})}.
\end{equation}

This motivates the following:

\begin{definition}
	\label{Rough}
	A closure operator $c_{U}: 2^U \longrightarrow 2^U$ on a set $U$ will be called a rough closure operator on $U$ if it satisfies
	\begin{equation}
		\label{RH}
		c_{U}(X)= U-c_{U}(U-c_{U}(X)) \ \ \ \ ~\mbox{for each} \ \  X  \subseteq U.
	\end{equation}
	The pair $(U,c_{U})$ is called a rough closure space. Dually, an interior operator $i_{U}: 2^U \longrightarrow 2^U$ on a set $U$ will be called a rough interior operator on $U$ if it satisfies
	\begin{equation}
		\label{RL}
		i_{U}(X)= U-i_{U}(U-i_{U}(X)) \ \ \ \ ~\mbox{for each} \ \  X  \subseteq U.
	\end{equation}
	The pair $(U,i_{U})$ is called a rough interior space.
\end{definition}

Notice that (\ref{RH}) is equivalent to $c_{U}(X)= i_{c_{U}}(c_{U}(X))$. This, together with (\ref{FixP}), leads to the following:

\begin{proposition}
	\label{Rcl}
	Let $c_{U}: 2^U \longrightarrow 2^U$ be a rough closure operator on a set $U$. Then it dual interior operator
	\begin{center}
		$i_{c_{U}} := 2^U \longrightarrow 2^U,~ X \longmapsto \ U-c_{U}(U-X)$
	\end{center}
	is a rough interior operator on $U$, and the topology
	\begin{eqnarray*}
		{\cal T}_{c_{U}} &=& \{X \subseteq U \mid c_{U}(U-X)=U-X \}  \nonumber\\
		&=& \{ X \subseteq U  \mid i_{c_{U}}(X)=X  \}
	\end{eqnarray*}
	associated with $c_{U}$ is clopen and therefore Alexandroff. Moreover, the set of $c_{U}(\{u\})$ is a basis for $U$. Indeed, it is unique minimal basis for $U$.
\end{proposition}

Consider the category ${\bf Cls}$ of closure spaces and continuous functions. We can define a second category ${\bf RCls}$ of all rough closure spaces and continuous functions, which is the subcategory of ${\bf Cls}$ with objects ordered pairs $(U,c_{U})$, where $U$ is a set and $c_{U}$ is a rough closure operator on $U$. Arrows $f : (U,c_{U}) \to (V,c_{V})$ are continuous functions; i.e., functions $f : U \to V$ such that
\begin{equation}
	\label{ClCont}
	f[c_{U}(X)]  \subseteq c_{V}(f[X]) \ \ \ \ \mbox{for all} \ X \subseteq U.
\end{equation}
or equivalently, ${c_{U}({f^{\leftarrow}(Y)}}) \subseteq f^{\leftarrow}(c_{V}(Y))$ for all $Y \subseteq V$. When thought of as a continuous function from $(U,c_{U})$ to $(V,c_{V})$, we will symbolize the function as $f : (U,c_{U}) \to (V,c_{V})$.

Consider two rough closure spaces $(U,c_{U})$ and $(V,c_{V})$, we have the two sets of arrows
\begin{center}
	${\rm hom}_{\bf Cls}((U,c_{U}),(V,c_{V})) \ \ \ \mbox{and} \ \ \ {\rm hom}_{\bf RCls}((U,c_{U}),(V,c_{V}))$
\end{center}
in the categories. As functions sets of functions, these two sets are the same. This shows that for each ordered pair $(U,c_{U})$, $(V,c_{V})$ of objects of ${\bf RCls}$,
\begin{equation}
	\label{RCls}
	{\rm hom}_{\bf Cls}((U,c_{U}),(V,c_{V}))={\rm hom}_{\bf RCls}((U,c_{U}),(V,c_{V})).
\end{equation}
This shows that
\begin{equation}
	\label{RClsCls}
	{\bf RCls} \ \mbox{is a full subcategory of} \ {\bf Cls}.
\end{equation}
Similarly, we define a category ${\bf RInt}$ of all rough interior spaces and continuous functions, which is the subcategory of ${\bf Int}$ with objects ordered pairs $(U,i_{U})$, where $U$ is a set and $i_{U}$ is a rough interior operator on $U$. Arrows $f : (U,i_{U}) \to (V,i_{V})$ are continuous functions; i.e., functions $f : U \to V$ such that
\begin{equation}
	\label{IntCont}
	f^{\leftarrow}(i_{V}(Y)) \subseteq  i_{U}(f^{\leftarrow}(Y)) \ \ \ \ \mbox{for all} \ Y \subseteq V.
\end{equation}
For each ordered pair $(U,i_{U})$, $(V,i_{V})$ of objects of ${\bf RInt}$, we have
\begin{equation}
	\label{RInt}
	{\rm hom}_{\bf Int}((U,i_{U}),(V,i_{V}))={\rm hom}_{\bf RInt}((U,i_{U}),(V,i_{V})).
\end{equation}
This shows that
\begin{equation}
	\label{RIntInt}
	{\bf RInt} \ \mbox{is a full subcategory of} \ {\bf Int}.
\end{equation}

\subsection{Equivalence of ${\bf AprS}$, ${\bf RCls}$ and ${\bf RInt}$}
Using (\ref{sym}) and Definition \ref{Rough}, we obtain

\begin{lemma}
	\label{RoughCI}
	Let $\langle U,t \rangle$ be an approximation space. Then
	\begin{enumerate}
		\item the upper approximation ${{\overline {apr}}_t}(X): 2^U \longrightarrow 2^U$ is a rough closure operator on $U$.
		\item the lower approximation ${{\underline {apr}}_t}(X): 2^U \longrightarrow 2^U$ is a rough interior operator on $U$.
	\end{enumerate}
\end{lemma}

Assume that $c: 2^U \longrightarrow 2^U$ is a rough closure operator on a set $U$.  It follows from (\ref{ClX}) and Proposition \ref{Rcl}, for each $u \in U$, $c(\{u\})$ is not only the smallest closed set including $u$, but also the smallest open neighborhood of $u$.
This, together with (\ref{min}), gives that for any $u,~v \in U$, $u \in c(\{v\})$ if and only if $v \in c(\{u\})$. It follows that $c(\{u\})$ and $c(\{u\})$ are either equal or they are disjoint. Therefore, the relation $t \subseteq U \times U$ defined by letting
\begin{center}
	$t = \bigcup\limits_{u \in U}(\{u\} \times c(\{u\}))$
\end{center}
is an equivalence relation on $U$. Moreover, $[u]_t=c(\{u\}))$ for each $u \in U$. Therefore, from (\ref{ClNbhd}) and (\ref{sym}), we have ${{\overline {apr}}_t}(X)=c(X)$ for each $X \subseteq U$. This proves the following:

\begin{proposition}
	\label{partition}
	Assume that $c: 2^U \longrightarrow 2^U$ is a rough closure operator on a set $U$. Let ${\cal T}_c$ be the topology associated with $c$, and let
	\begin{equation}
		\label{equiv}
		t = \bigcup\limits_{u \in U}(\{u\} \times c(\{u\})).
	\end{equation}
	Then $t$ is an equivalence relation on $U$, and
	\begin{center}
		${{\overline {apr}}_t}(X)=c(X) \ \ \ \ ~\mbox{for each} \ \  X  \subseteq U$.
	\end{center}
\end{proposition}

According to Proposition \ref{RelTop}, Lemma \ref{RoughCI}, (\ref{ClCont}) and Proposition \ref{partition}, approximation spaces on a set $U$ are in bijective correspondence with rough closure operators on the set $U$; moreover, the relation-preserving functions between approximation spaces are precisely the continuous functions between the corresponding rough closure.

According to Proposition \ref{RelTop} and Lemma \ref{RoughCI}, we can define a functor
\begin{center}
	$F : {\bf AprS} \to {\bf RCls}$
\end{center}
by setting $F(\langle U,t \rangle)=(U,{{\overline {apr}}_t}: 2^U \longrightarrow 2^U$) for any object $\langle U,t \rangle$ of ${\bf AprS}$, and for $f : \langle U,t \rangle \to \langle V,s \rangle$ setting
\begin{center}
	$F(f : \langle U,t \rangle \to \langle V,s \rangle)=f : (U,{{\overline {apr}}_t}: 2^U \longrightarrow 2^U) \to (V,{{\overline {apr}}_s}: 2^V \longrightarrow 2^V)$.
\end{center}
According to Lemma \ref{RoughCI}, (\ref{ClCont}) and Proposition \ref{partition}, we can define a functor
\begin{center}
	$F' : {\bf RCls} \to {\bf AprS}$
\end{center}
by setting $F'((U,c_{U}))=\langle U,t \rangle$, where
\begin{center}
	$t = \bigcup\limits_{u \in U}(\{u\} \times c_{U}(\{u\}))$,
\end{center}
for any object $(U,c_{U})$ of ${\bf RCls}$, and for $f : (U,c_{U}) \to (V,c_{V})$ setting
\begin{center}
	$F'(f : (U,c_{U}) \to (V,c_{V}))=f : \langle U,t \rangle \to \langle V,s \rangle$,
\end{center}
where $s = \bigcup\limits_{v \in V}(\{v\} \times c_{V}(\{v\}))$. It is easy to check that
\begin{equation}
	\label{equiv}
	F' \circ F = {\bf 1}_{\bf AprS}\;\;\mbox{\rm and}\;\; F\circ F' = {\bf 1}_{\bf RCls}.
\end{equation}
This proves that the two categories ${\bf AprS}$ and ${\bf RCls}$ are isomorphic as objects of ${\bf Cat}$.

Similarly, according to (\ref{dualC}) and Proposition \ref{Rcl}, it is easy to check that the functor
\begin{center}
	$G : {\bf RCls} \to {\bf RInt}$
\end{center}
defined by setting $G((U,c_{U}))=(U,i_{c_{U}})$, where
\begin{center}
	$i_{c_{U}} := 2^U \longrightarrow 2^U :~ X   \ \longmapsto \  U-c(U-X)$,
\end{center}
and for $f : (U,c_{U}) \to (V,c_{V})$ setting
\begin{center}
	$G(f : (U,c_{U}) \to (V,c_{V}))=f : (U,i_{c_{U}}) \to (V,i_{c_{V}}))$,
\end{center}
in the category ${\bf Cat}$ is an isomorphism. Consequently, the three categories ${\bf AprS}$, ${\bf RCls}$ and ${\bf RInt}$ are isomorphic as objects of ${\bf Cat}$.

\subsection{The category of information systems}
The category of information systems ${\bf IS}$ has information systems as its objects and O-A-D homomorphisms as its arrows. We say that an O-A-D homomorphism is {\em non-expansive\/} if its attribute mapping is onto. To establish the relationship between ${\bf IS}$ and  ${\bf AprS}$, we consider a subcategory ${\bf NeIS}$ whose objects are information systems and arrows are non-expansive O-A-D homomorphisms. Then, it is easy to see that, if $h=(h_O,h_A,h_D)$ is a non-expansive O-A-D homomorphism between information systems $(U,A_1, D_1)$ and $(V,A_2,D_2)$, then $h_O$ is a relation-preserving function between approximation space $\langle U,t_{A_1}\rangle$ and $\langle V,t_{A_2}\rangle$ by using (\ref{oad}). Hence, we can define a functor
\begin{center}
	$H : {\bf NeIS}\to {\bf AprS}$
\end{center}
by setting $H((U,A,D))=\langle U,t_A\rangle$ and \[H(h:(U,A_1, D_1)\to (V,A_2,D_2))=h_O:\langle U,t_{A_1}\rangle\to\langle V,t_{A_2}\rangle.\]

On the other hand, let $f:\langle U,t\rangle\to \langle V,s\rangle$ be a relation-preserving function. Then, we can define a non-expansive O-A-D homomorphism $h_f=(h_O,h_A,h_D): (U,\{a_t\}, U/t)\to (V,\{a_s\}, V/s)$ by setting $h_O=f$, $h_A(a_t)=a_s$, and $h_D([u]_t)=[f(u)]_s$ for any $[u]_t\in U/t$. It is obvious that $h_f$ is non-expansive and satisfies the homomorphic condition (\ref{oad}). Thus, we can define a functor
\begin{center}
	$H' : {\bf AprS}\to {\bf NeIS}$
\end{center}
by setting
$H'(\langle U,t\rangle)=(U,\{a_t\}, U/t)$  and \[H'(f:\langle U,t\rangle\to \langle V,s\rangle)=h_f:(U,\{a_t\}, U/t)\to (V,\{a_s\}, V/s).\]
We can see that \[H\circ H'={\bf 1}_{\bf AprS}\]
but $H'\circ H={\bf 1}_{\bf NeIS}$ does not hold. This means that an information system is more informative than the finest approximation space derived from it. Indeed, we can usually induce the same approximation space from several different information systems by using different sets of attributes, and this is the main idea of attribute reduction in rough set theory.

\section{Related Works}
While rough set theory has been extensively studied from diverse perspectives, there were a few works on its categorical aspect. In \cite{Borzooei}, three categories of approximations, denoted by $\underline{\bf Apr}{\bf S}$, $\overline{\bf Apr}{\bf S}$, and ${\bf AprS}$, are defined. Objects of the three categories, as well as our definition of ${\bf AprS}$,  are all approximation spaces. However, the require arrows to preserve not only the underlying equivalence relation but also lower and/or upper approximations. These arrows are called lower/upper transformations although they are simply morphisms instead of natural transformations in the sense of category theory.  More precisely, let $\langle U,t\rangle$ and $\langle V,s\rangle$ be two approximation spaces. Then, a function $f:U\to V$ is  called a lower and upper natural transformation if it satisfies, for any $X\subseteq U$,
\[\underline{apr}_s(f[X])=f[\underline{apr}_t(X)]\] and
\[\overline{apr}_s(f[X])=f[\overline{apr}_t(X)]\]respectively. In addition, $f$ is simply a natural transformation if it is both a lower and upper natural transformation. It is then proved that a lower natural transformation preserves equivalence classes (\cite{Borzooei},Proposition 2.4). On the other hand, they also proved that a mapping is an upper natural transformation if and only if it preserves equivalence classes (\cite{Borzooei}, Theorem 3.4). This indicates that a lower natural transformation is automatically an upper natural transformation, although it is generally not the case. Because a function preserving equivalence classes is necessarily relation-preserving (but not conversely), the categories introduced in \cite{Borzooei} imposes a much stricter restriction on morphisms than ours. As a result, all three categories in \cite{Borzooei} are subcategories of ${\bf AprS}$ (or equivalently, ${\bf RCls}$ or ${\bf RInt}$) in this paper.

In fact, the morphisms defined in \cite{Borzooei} are simply instances of modal homomorphisms commonly used in algebraic modal logic~\cite{blak}. In the algebraic interpretation of modal logic, well-formed formulas are interpreted as elements of {\em Boolean algebras with operators\/}(BAOs). A BAO is an algebra $\mathfrak{A}=(A,+,-,0,\vartriangle)_{\vartriangle\in\tau}$, where $(A,+,-,0)$ is a Boolean algebra and $\tau$ is a set of modal operators such that each $\vartriangle\in\tau$ has a arity $\rho(\vartriangle)>0$, such that the following two conditions are satisfied
\begin{enumerate}
	\item normality: $\vartriangle\!\!(a_1,\cdots,a_{\rho(\vartriangle)})=0$ if $a_i=0$ for some $0<i\leq\rho(\vartriangle)$.
	\item additivity: for any $0<i\leq\rho(\vartriangle)$, \begin{eqnarray}\vartriangle\!\!(a_1,\cdots,a_i+a_i',\cdots,a_{\rho(\vartriangle)})&=&\\
\vartriangle\!\!(a_1,\cdots,a_i,\cdots,a_{\rho(\vartriangle)})&+&\vartriangle\!\!(a_1,\cdots,a_i',\cdots,a_{\rho(\vartriangle)}).\end{eqnarray}
\end{enumerate}
It is well-known that lower and upper approximations are set-theoretic counterparts of modal operators $\Box$ and $\Diamond$ in S5 modal logic. Hence, in the current context, we need to consider a special BAO with only one modal operator $\diamond$ with arity 1. In such a BAO, the normality and additivity conditions reduce to $\diamond 0=0$ and $\diamond(a+a')=\diamond a+\diamond a'$ respectively. In addition, extra algebraic axioms for S5 logic are T: $a+\diamond a=\diamond a$, 4: $\diamond\diamond a=\diamond a$, and 5: $-\!\diamond a=\diamond(-\!\diamond a)$. It is easy to see that the rough closure space $(2^U,\cup,-,\emptyset, c_U)$ is an instance of this special kind of S5 BAO. Let $\mathfrak{A}=(A,+,-,0,\diamond)$ and $\mathfrak{A}'=(A',+',-',0',\diamond')$ be two BAOs. Then, a function $f:A\to A'$ is called a modal homomorphism~\cite{blak} if it satisfies
\[f(\diamond a)=\diamond' f(a).\] Hence, regarding $(2^U,\cup,-,\emptyset, \overline{apr}_t)$ and $(2^V,\cup,-,\emptyset, \overline{apr}_s)$ as concrete instances of S5 BAOs and a function $f:U\to V$ as a mapping $f:2^U\longrightarrow 2^V$, the upper natural transformation introduced in \cite{Borzooei} is indeed a modal homomorphism. In addition, if the upper natural transformation $f:2^U\longrightarrow 2^V$ is also a homomorphism between the two Boolean algebras $(2^U,\cup,-,\emptyset)$ and $(2^V,\cup,-,\emptyset)$ (in particular, $f[-X]=-f[X]$ for any $X\subseteq U$), then it is also a lower natural transformation.

Unlike categories defined in \cite{Borzooei} and this paper, there are also categories of rough sets whose objects are not only approximation spaces. The earliest categorical analysis of rough sets is the category {\footnotesize\bf ROUGH} defined in \cite{Ban}. Objects of {\footnotesize\bf ROUGH} are triples $\langle R,t,X\rangle$, where $\langle R,t\rangle$ is an approximation space and $X\subseteq U$. Let $\overline{\mathcal X}_t$ and $\underline{\mathcal X}_t$ denote the quotient sets of $\overline{apr}_t(X)$ and $\underline{apr}_t(X)$ respectively. Then, an arrow $f:\langle R,t,X\rangle\to \langle V,s,Y\rangle$ in {\footnotesize\bf ROUGH} is a map $f:\overline{\mathcal X}_t\to\overline{\mathcal Y}_s$ such that $f[\underline{\mathcal X}_t]\subseteq\underline{\mathcal Y}_s$. Thus, arrows of {\footnotesize\bf ROUGH} must preserve lower approximations.

Yet another category of rough sets is based on the algebraic interpretation of rough sets~\cite{Iwin}. A pair $(U,{\bf B})$ is called a {\em rough universe} in \cite{Iwin}, where $U$ is the domain and {\bf B} is a subalgebra of the power set Boolean algebra $(2^U,\cup,-,\emptyset)$. Any pair $(X_1,X_2)$ such that $X_1, X_2\in {\mathbf B}$ and $X_1\subseteq X_2$ is an I-rough set of $(U,{\bf B})$. Then, the category {\footnotesize\bf RSC} has all I-rough sets as its objects and an arrow $f:(X_1,X_2)\to(Y_1,Y_2)$ is simply a function $f:X_2\to Y_2$ such that $f[X_1]\subseteq Y_1$~\cite{More}. While {\footnotesize\bf ROUGH} and  {\footnotesize\bf RSC} looks different at first glance, they are shown to be equivalent in \cite{More}.

A subcategory of {\footnotesize\bf ROUGH}, called $\xi$-{\footnotesize\bf ROUGH}, is defined to have same objects as {\footnotesize\bf ROUGH}, but its arrows must satisfy the additional condition of preserving the boundary region. That is, $f:\langle R,t,X\rangle\to \langle V,s,Y\rangle$ is an arrow of $\xi$-{\footnotesize\bf ROUGH} if it is an arrow of {\footnotesize\bf ROUGH} satisfying $f[\overline{\mathcal X}_t-\underline{\mathcal X}_t]\subseteq
(\overline{\mathcal Y}_s-\underline{\mathcal Y}_s)$. Analogously, $\xi$-{\footnotesize\bf RSC} is the subcategory of {\footnotesize\bf RSC} with the same collection of objects and its arrows $f:(X_1,X_2)\to(Y_1,Y_2)$ are {\footnotesize\bf RSC} arrows satisfying $f[X_2-X_1]\subseteq(Y_2-Y_1)$. It is then shown that $\xi$-{\footnotesize\bf ROUGH} and $\xi$-{\footnotesize\bf RSC} are equivalent and both are equivalent to {\bf Set}$^2$, whose objects are pairs of sets $(X_1,X_2)$ and arrows are pairs of functions $(f,g):(X_1,X_2)\to(Y_1,Y_2)$ where $f:X_1\to Y_1$ and $g:X_2\to Y_2$. The result shows that the role of approximation spaces has become hardly visible in the categories of algebraic rough sets. By contrast, our categories of approximation spaces, rough closure spaces, and rough interior spaces arise from a topological interpretation of rough sets. Hence, approximation spaces and  continuous functions between them play the major role in such categories. In some sense, this means that categories based on I-rough sets somewhat lose structural information behind the construction of rough approximations.

On the other hand, the advantage of using I-rough sets is that more general categorical construction is possible. In \cite{More}, by changing the base from {\bf Set} to an arbitrary topos, a natural generalization of {\footnotesize\bf RSC} and $\xi$-{\footnotesize\bf RSC} is obtained. Recalling that a topos, as a category-theoretic abstraction of {\bf Set}, is a category $\mathscr{C}$ that has all finite limits, a subobject classifier, and all exponentials~\cite{Awodey}. Then, the generalization of {\footnotesize\bf RSC} (resp.\ $\xi$-{\footnotesize\bf RSC})  to a parameterized class of categories {\footnotesize\bf RSC}$(\mathscr{C})$ (resp.\ $\xi$-{\footnotesize\bf RSC}$(\mathscr{C})$) is proposed and its topos-theoretic properties are explored in \cite{More}. Objects of {\footnotesize\bf RSC}$(\mathscr{C})$ and $\xi$-{\footnotesize\bf RSC}$(\mathscr{C})$ are both triples $(X,Y,m)$, where $X$ and $Y$ are objects of $\mathscr{C}$ and $m:X\to Y$ is a monic arrow in $\mathscr{C}$. An arrow of {\footnotesize\bf RSC}$(\mathscr{C})$ is a pair $(f,g):(X,Y,m)\to (X',Y',m')$ where $f:X\to X'$ and $g:Y\to Y'$ are arrows in $\mathscr{C}$ such that $g\circ m=m'\circ f$, and it is also a $\xi$-{\footnotesize\bf RSC}$(\mathscr{C})$ arrow if there exist objects $(\neg X,Y,\neg m)$ and $(\neg X',Y',\neg m')$ and arrow $f^-:\neg X\to\neg X'$ in $\mathscr{C}$ such that $\neg m'\circ f^-=g\circ\neg m$. By instantiating $\mathscr{C}$ to a particular topos {\bf M-Set} (the category of actions of a monoid {\bf M} on sets, see \cite{Gold}), a possible application of {\footnotesize\bf RSC}$(\mathscr{C})$ to monoid actions on rough sets is suggested in \cite{More}.

\section{Conclusion}
With the notions of  rough closure operator and rough interior operator together with the functor, we obtain the equivalence of the category of all rough interior spaces and continuous functions, the category of rough closure spaces and continuous  functions, the category of approximation spaces and relation-preserving functions. Our approach using information system to characterize the category of approximation space and its subcategory gives rise to a deeper understanding of the interplay among rough set theory, information system and category theory.  It gives us not only a better understanding of the rough set theory structures  in the context of category theory but also the relationships of different categories arising from some topological tools. By using the concepts of categories, one can further derive useful properties and applications of rough set theory. More precisely, category theory can be used to show how mathematical structures of a given type can be transformed into one another by functions that preserve some aspect of their structures; it provides a uniform language for speaking of various mathematical or topological structures and the mappings within those types. The theory of abstract categories, however, totally ignores the sets, operations, relations and axioms that specify the structure of objects in question. It just provides a language in which one can talk about how mappings preserve certain structure behaviors without a good understanding of the relationship of various properties or implications of what have been proved. For that reason, more technical insights and knowledge will help explore further theory and applications. We envision to work out more results along this line of research.

\section*{Acknowledgement}
This work was partially supported by the Ministry of Science and Technology (TAIWAN) with grant numbers 109-2221-E-150-028 (for Y.R. Syau) and 110-2221-E-001-022-MY3 (for C.J. Liau).

\begin{small}

\end{small}

\end{document}